\documentclass[twocolumn,10pt]{article}
\usepackage[margin=1in]{geometry}
\usepackage{cite}
\usepackage{amsmath,amssymb,amsfonts}
\usepackage{algorithmic}
\usepackage{graphicx}
\usepackage{textcomp}
\usepackage{hyperref}
\usepackage{etoolbox}
\usepackage{caption}
\usepackage{amsmath,graphicx}
\usepackage{booktabs}
\usepackage{float}
\usepackage{tabularx}

\title{LMM-IQA: Image Quality Assessment for Low-Dose CT Imaging}

\author{
Kagan Celik$^{1}$, Mehmet Ozan Unal$^{1}$, Metin Ertas$^{2}$, Isa Yildirim$^{1}$\\[4pt]
\small $^{1}$Department of Electronics and Communication Engineering, Istanbul Technical University, Istanbul, Turkiye\\
\small \textbf{Corresponding author:} \texttt{celikk23@itu.edu.tr}
}

\begin{document}
\maketitle
\begin{abstract}
Low-dose computed tomography (CT) represents a significant improvement in patient safety through lower radiation doses, but the increase of noise and blur and contrast loss can all diminish the diagnostic quality of the images. Therefore, consistency and robustness in image quality assessment becomes an essential aspect for clinical applications. 
In this study, we propose an LLM-based quality assessment system that generates both numerical scores and textual descriptions of degradations such as noise, blur, and contrast loss. Furthermore, various inference strategies — from the zero-shot approach to metadata integration and error feedback — are systematically examined, demonstrating the progressive contribution of each method to the overall performance. The resultant assessments yield not only highly correlated scores, but also interpretable output, thereby adding value to clinical workflows.
The source codes of our study are available at \textbf{\url{https://github.com/itu-biai/lmms_ldct_iqa}}.
  
\end{abstract}

\begin{figure*}[htbp]
    \centering
    \includegraphics[width=0.95\textwidth]{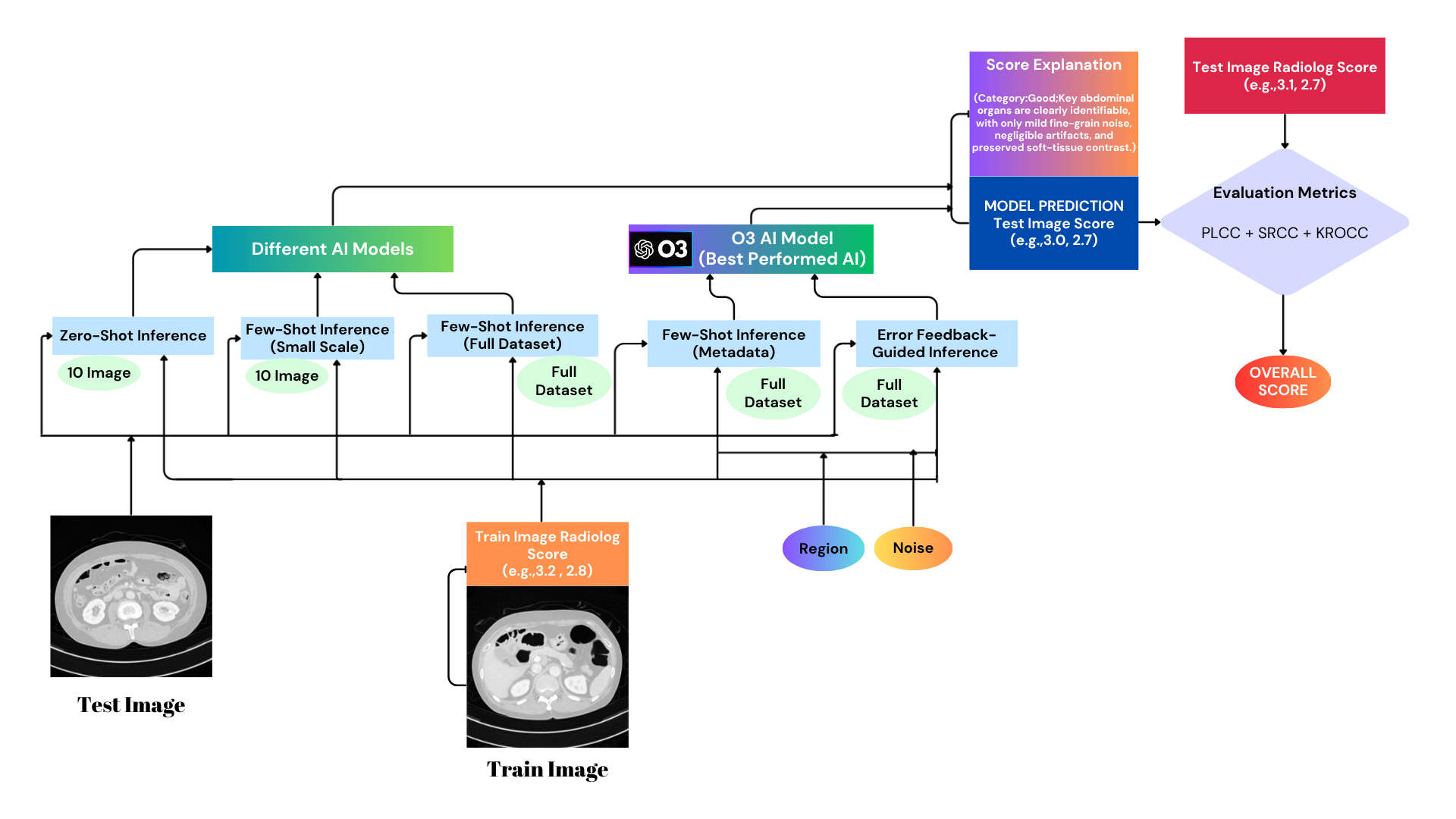}
    \caption{Flowchart of the LMM-IQA Methodology}
    \label{fig:flowchart}
\end{figure*}


\section{Introduction}
Computed tomography (CT) is one of the most commonly employed imaging modalities in contemporary medicine and it is essential to diagnostic accuracy and patient management. However, the hazards associated with ionizing radiation represent a serious challenge to patient safety. Thus, the development and utilization of LDCT protocols has increased \cite{lee2025low}. LDCT provides a value in that it utilizes minimal radiation doses, but it can be associated with changes in image quality, noise and artifacts, which have a very real impact on radiologists' diagnostic outcomes. Therefore, the dependable and high-quality evaluation of LDCT must be guaranteed for accurate clinical judgments.

Image quality relies on subjective assessment by radiologists. While this is the data gold-standard approach, it is beset by disadvantages in the guise of interobserver variability, time-consuming workload, and time expense. Automatically applied Image Quality Assessment (IQA) methods therefore have arisen as an extremely targeted research activity in recent years~\cite{bosse2017deep,zhu2020metaiqa,yang2022maniqa,madhusudana2022image,talebi2018nima}.
PSNR and SSIM, used in early full-reference approaches, often failed to capture diagnostic relevance, which led to the development of no-reference (NR) IQA methods.

Recently, in addition to these full-reference metrics, no-reference (NR) IQA approaches, and particularly deep learning-based methods, have also made significant progress. Various deep learning methods show high correlation with radiologists' scores \cite{zhu2020metaiqa,yang2022maniqa,madhusudana2022image,talebi2018nima}. However, these approaches require large and balanced datasets, which often results in generalization issues when imaging modalities change, thereby necessitating retraining.

In recent years, a significant paradigm shift in artificial intelligence has taken place. Large language models (LLM) and their multimodal variants achieve unique performance not only for language processing approaches but also for more complicated visual-text-based problems \cite{wang2024comprehensive}. Because of their unique abilities in text generation, logical inference, and contextual understanding, LLMs can replace traditional deep learning approaches with a more flexible and generalized solution. Multimodal LLMs (LMMs), thanks to their capacity to evaluate images alongside linguistic explanations, not only produce a score but also provide human-like justifications \cite{you2024depicting,youenhancing,chen2024grounding}. These features hold significant potential in terms of both explainability and relevance to clinical contexts in IQA tasks \cite{zhu2024adaptive,varga2025comparative,wu2023q,you2025teaching}.

While several attempts at LMM-based IQA exist in the literature~\cite{chen2024iqagpt,wang2024comprehensive,you2024depicting,youenhancing}, no research has yet explored such a comprehensive dataset and multi-model comparison in the context of LDCT. Our study is based on a large-scale dataset with high confidence (ICC=0.96) consisting of 1000 training and 300 test images provided by the MICCAI 2023 LDCT-IQA Challenge \cite{lee2025low}. 

In this study, we systematically investigate LMMs for image quality assessment on a large-scale LDCT dataset, introducing key innovations such as a no-reference evaluation setup, zero-shot and few-shot inference testing, an Error Feedback–Guided Inference strategy, and metadata-based explainability enhancement. Furthermore, the comparative evaluation of multiple
AI models strengthened the reliability of the obtained results. In summary, these advancements introduce a flexible, training-free few-shot inference framework that is readily adaptable to clinical contexts and extends beyond conventional deep learning–based IQA methods.

\section{Methodology}
\subsection{Dataset}
This study used the open-access dataset generated as part of the MICCAI 2023 Low-Dose Computed Tomography Perceptual Image Quality Assessment (LDCT-IQA) Challenge \cite{lee2025low}. The dataset consists of 1,000 training and 300 test images. The images were sourced from the Mayo Clinic (USA) and the National Cancer Center (South Korea) and were obtained from different populations.
A physics-based artifact addition pipeline was applied to the images to reflect low-dose conditions by introducing varying levels of Poisson noise and streak artifacts. Radiologists scored all images on a 0–4 Likert scale (0: very poor, 4: excellent). For each image, the average of the scores provided by 5–6 radiologists was used to create a ground-truth. These annotations were found to be highly reliable, with an intraclass correlation coefficient (ICC) value of 0.96 in the test set.

\subsection{Evaluated Large Language and Multimodal Models}
The study tested large-scale language and multimodal models from different vendors:
OpenAI: GPT-o3, GPT-4o, GPT-4o-mini; Google: Gemini 2.5 Pro; Meta: Llama-4 Maverick; Qwen: VL Max; Anthropic: Claude Sonnet-4; xAI: Grok 2 Vision; Google DeepMind: Gemma 3.
These models were tested in both zero-shot and few-shot scenarios, consistent with recent multimodal IQA benchmarks \cite{varga2025comparative,you2024depicting,wu2023q,you2025teaching}.

\subsection{Proposed LMM-Based IQA Frameworks}
Fig.~\ref{fig:flowchart} illustrates the general operation of the different approaches in the proposed LMM-IQA methods. The model follows a process that includes zero-shot, few-shot, metadata-aided, and error-feedback-guided inference steps. In these stages, the contextual accuracy of the predictions is increased by adding external region and noise information. Ultimately, the model produces both numerical quality scores and descriptive text, demonstrating a gradual increase in correlation with radiologist assessments.

\subsubsection{Zero-Shot Inference}
Models scored test images without any sample images. Scoring was done with 10 test images. Result of clustering were saved to a separate file for analysis.

\subsubsection{Few-Shot Inference (Small-Scale)}
10 training images and their related radiolog scores were given to the model, and then 10 test images were requested as scored from model, consistent with few-shot setups in prior work \cite{you2025teaching}.

\subsubsection{Few-Shot Inference (Full Dataset)}
In this setup, the models were evaluated on the entire dataset (1000 training and 300 test images) using a training-free few-shot inference paradigm. Rather than parameter optimization, a limited number of training examples were provided in-context to guide the model’s reasoning. For each inference step, 10 representative examples were included in the prompt, while 34 examples were used in the final tests to further stabilize performance. Since there were 300 test images, the models produced one prediction per test image across 300 inference steps. In addition to numerical scores, the models generated brief textual explanations describing the criteria used for scoring, similar to the descriptive IQA frameworks proposed in \cite{youenhancing}.

\subsubsection{Error Feedback–Guided Inference}
In this study, Error Feedback–Guided Inference approach was applied to model evaluation. 
In the first stage, the model evaluated several randomly selected training images with the corresponding noise level ($n_i$) that was generated from noise estimator 
and the true score ($y_i$) was determined by the radiologist. 
The absolute error, 
\begin{equation}
e_i = |y_i - \hat{y}_i|,
\end{equation}
was calculated as the difference between the model's prediction ($\hat{y}_i$) and the true score ($y_i$). 
These error values were provided as feedback to the model for each sample and were used as guidance for subsequent test samples. 
Thus, the model attempted to minimize the error in its new predictions by considering the errors in its past predictions. 
The reward–punishment loop can be defined similarly to the expected reward function in the classical Reinforcement Learning (RL) literature. 
The model's performance metric is to minimize the total error: 
\begin{equation}
J(\theta) = \arg\min_{\theta} \mathbb{E}_{(x, y) \sim D}[\,|y - f_\theta(x)|\,]
\end{equation}
where $(x, y) \sim D$ denotes the data sampled from the training distribution, 
and $f_\theta(x)$ represents the model’s prediction parameterized by $\theta$. 
The resulting $e_i$ errors at each new prediction step are passed on to the model as a negative penalty. 
Unlike classical supervised learning, which relies solely on assigning correct labels, this method improves the model's decision-making process by converting error magnitude into information \cite{wu2023q}.

\subsubsection{Metadata Integration for Context-Aware Scoring}
Anatomical \textit{Region} labels were added to each image to increase the model’s scoring accuracy, enabling it to better understand structural differences and make more accurate comparisons between similar anatomical areas.  
For this purpose, the LDCT images were provided to the LLMs to determine the corresponding anatomical \textit{Region} (e.g., abdomen, kidney, etc.).
Both training and testing images were assigned with these labels and used as input in LMM few-shot scenarios. 
This aimed to allow the model to more accurately capture quality differences between the same anatomical regions.

\textit{Noise} level is one of the most critical factor determining the quality of CT images. Therefore, Poisson-Gaussian-based noise estimation was performed for each image. No reference-noise metadata were estimated quickly and accurately using the Fast Blind Image Denoiser (FBI-Denoiser) method \cite{byun2021fbi}.
For example, values such as 0.003 or 0.005 were added to each image as noise metadata, and '\textit{Region} + \textit{Noise}' were incorporated into the LMM input to produce more consistent predictions with radiologist.

Consequently, the addition of Region information allowed the model to better understand structural differences, contributing to more accurate comparisons between similar anatomical areas. 
The addition of Noise information allowed it to account for noise variations inherent in low-dose CT and helped produce predictions more consistent with radiologist scores.

\subsection{Evaluation Metrics}
Three correlation coefficients were used to evaluate model performance, measuring the relationship between predicted scores and radiologist scores: PLCC, SROCC, and KROCC. These metrics assess model reliability by capturing both linear correlation and rank-based similarities, in line with prior LDCT-IQA definitions \cite{lee2025low}.

\textbf{PLCC (Pearson Linear Correlation Coefficient)} measures the linear correlation between model predictions and radiologist scores. It is defined mathematically as follows:

\begin{equation}
\mathrm{PLCC} = \frac{\sum (s_i - \mu)(\hat{s}_i - \hat{\mu})}{\sqrt{\sum (s_i - \mu)^2 \sum (\hat{s}_i - \hat{\mu})^2}},
\end{equation}

where $s_i$ represents the score given by the radiologist, and $\hat{s}_i$ represents the score predicted by the model. The terms $\mu$ and $\hat{\mu}$ denote the averages of the radiologist and model scores, respectively.

\textbf{SROCC (Spearman Rank Order Correlation Coefficient)} measures the rank order similarity of scores and is used to evaluate the relative consistency between rankings:

\begin{equation}
\mathrm{SROCC} = 1 - \frac{6 \sum (R_i - \hat{R}_i)^2}{n(n^2 - 1)},
\end{equation}

where $R_i$ and $\hat{R}_i$ represent the ranks of the actual and predicted scores, respectively, and $n$ denotes the total number of images.

\textbf{KROCC (Kendall Rank Correlation Coefficient)} is calculated based on the difference between the concordant pairs ($M_c$) and discordant ($M_d$) pairs of two rankings:

\begin{equation}
\mathrm{KROCC} = \frac{M_c - M_d}{\sqrt{(M_c + M_d + T_s)(M_c + M_d + T_t)}},
\end{equation}

where $T_s$ and $T_t$ represent the terms accounting for ties in the rankings.

In the final evaluation of the results, these three metrics were considered together, and their sum was reported as the \emph{Overall Score}. Thus, model performance was comprehensively measured by considering both linear and rank-based similarities simultaneously. \cite{lee2025low}.

\section{Experimental Results}
\subsection{Zero-Shot Inference Results}
All AI models performed poorly in the zero-shot scenario as seen in Table~\ref{tab:zero_shot_results}. Although the GPT-4o-mini and O3 models produced relatively better results, a strong correlation was not established with radiologist scores. Therefore, given the low scores obtained even with 10 test images, a zero-shot experiment was deemed unnecessary on all test images. This finding demonstrates that the zero-shot approach might not be a good candidate for medical IQA.
\begin{table}[H]
    \centering
    \caption{Zero-shot inference results of different AI models}
    \label{tab:zero_shot_results}
    \footnotesize
    \renewcommand{\arraystretch}{1.1}
    \resizebox{\columnwidth}{!}{%
        \begin{tabular}{lcccc}
            \hline
            Model & PLCC & SROCC & KROCC & Overall Score \\
            \hline
            GPT-4o Scores         & 0.5015 & 0.2730 & 0.1673 & 0.9418 \\
            Claude-Sonnet4 Scores & 0.4221 & 0.3597 & 0.2948 & 1.0766 \\
            Gemma3 Scores         & 0.4941 & 0.4912 & 0.4138 & 1.3990 \\
            Grok2 Scores          & 0.6179 & 0.5704 & 0.4800 & 1.6683 \\
            Qwen Scores           & 0.5796 & 0.6174 & 0.4945 & 1.6915 \\
            Gemini Scores         & 0.6729 & 0.6863 & 0.5196 & 1.8788 \\
            Meta-Llama4 Scores    & 0.6595 & 0.6743 & 0.5700 & 1.9038 \\
            O3 Scores             & 0.7521 & 0.6796 & 0.5512 & 1.9828 \\
            GPT-4o-mini Scores    & 0.7325 & 0.7082 & 0.5457 & \textbf{1.9864} \\
            \hline
        \end{tabular}
    }
\end{table}

\subsection{Few-Shot Inference Results (Small Scale - 10 Images)}
The Small Scale Few Shot test results are shown in Table~\ref{tab:fewshot_small}.In few-shot experiments with a limited number of images (10 training + 10 test), the model performance improved significantly. The top-performing models were O3, GPT-4o, and Qwen, with Overall Scores in the range of approximately 2.60–2.67. The medium-performing models included GPT-4o-mini, Gemini, and Llama4, achieving Overall Scores between 2.34–2.54. Finally, the low-performing models were Gemma-3, Grok-2, and Claude Sonnet-4. This result demonstrates that a small number of samples allows the models to make predictions closer to radiologist scores. 
\begin{table}[H]
    \centering
    \caption{Few-shot inference results (Small Scale)}
    \label{tab:fewshot_small}
    \footnotesize
    \renewcommand{\arraystretch}{1.1}
    \resizebox{\columnwidth}{!}{%
        \begin{tabular}{lcccc}
            \hline
            Model & PLCC & SROCC & KROCC & Overall Score \\
            \hline
            Gemma3 Scores         & 0.2848 & 0.4431 & 0.3568 & 1.0847 \\
            Grok2 Scores          & 0.6760 & 0.6896 & 0.5552 & 1.9208 \\
            Claude Sonnet4 Scores & 0.8170 & 0.8006 & 0.6838 & 2.3014 \\
            Meta Llama4 Scores    & 0.8123 & 0.8115 & 0.7230 & 2.3468 \\
            Gemini Scores         & 0.8950 & 0.8329 & 0.7389 & 2.4669 \\
            GPT-4o-mini Scores    & 0.9222 & 0.8697 & 0.7483 & 2.5402 \\
            Qwen Scores           & 0.9095 & 0.8867 & 0.8040 & 2.6003 \\
            GPT-4o Scores         & 0.9606 & 0.8785 & 0.8015 & 2.6405 \\
            O3 Scores             & 0.9630 & 0.8927 & 0.8142 & \textbf{2.6698} \\
            \hline
        \end{tabular}
    }
\end{table}

\subsection{Few-Shot Inference Results (Full Dataset – 1000 Training / 300 Test)}
In experiments using the full dataset, the models were tested with 300 inference steps, with 10 training images per inference step and 1 test image as the output. However, performance decreased compared to the smaller-scale experiments. The initial test results are provided in Table~\ref{tab:fewshot_full}.The performance degradation stems from the model's limited contextual capacity and generalization ability. While the model learned relationships easily with small data because the examples were similar, the increased diversity across the entire dataset (different regions and noise levels) disrupted this consistency. Because LMMs do not train, they only infer short contexts; this reduces the representativeness of the examples in the large dataset, leading to a decrease in correlation.
The top-performing models were O3, GPT-4o, and Gemini, with Overall Scores in the range of approximately 1.85–1.88. The medium-performing models included GPT-4o-mini and Llama4, achieving Overall Scores between 1.02–1.26. Finally, the low-performing model was Gemma-3. These findings suggest that LMM models, which achieve high correlation in small-scale tests, experience performance degradation on larger and more diverse datasets.  
\begin{table}[H]
    \centering
    \caption{Few-shot inference results (Full Dataset)}
    \label{tab:fewshot_full}
    \footnotesize
    \renewcommand{\arraystretch}{1.1}
    \resizebox{\columnwidth}{!}{%
        \begin{tabular}{lcccc}
            \hline
            Model & PLCC & SROCC & KROCC & Overall Score \\
            \hline
            Gemma3 Scores         & 0.0677 & 0.0400 & 0.0300 & 0.1378 \\
            GPT-4o-mini Scores    & 0.3808 & 0.3664 & 0.2733 & 1.0205 \\
            Meta-Llama4 Scores    & 0.4765 & 0.4591 & 0.3261 & 1.2616 \\
            GPT-4o Scores         & 0.6781 & 0.6709 & 0.4996 & 1.8486 \\
            Gemini Scores         & 0.6758 & 0.6906 & 0.5093 & 1.8757 \\
            O3 Scores             & 0.6863 & 0.6854 & 0.5127 & \textbf{1.8844} \\
            \hline
        \end{tabular}
    }
\end{table}

\subsection{Few-Shot Inference Results with Metadata}
For the performance degradation reasons mentioned in full dataset test, the methods mentioned in the metadata section were used to generate performance improvements, and these improvements are shown step by step in Table~\ref{tab:fewshot_metadata} . Furthermore, in the final tests conducted to maximize performance, a maximum of 34 training images were given to the O3 model instead of 10 training images per inference step to allow the system to learn better. Tests were conducted using the entire dataset.

For both cost-effectiveness and optimal results achieved from previous experiments, test scenarios were developed using the O3 AI model in the optimization efforts. The final test results are shown in Table~\ref{tab:fewshot_metadata}.

\begin{table}[H]
    \centering
    \caption{O3 model few-shot inference results with different approaches}
    \label{tab:fewshot_metadata}
    \footnotesize
    \renewcommand{\arraystretch}{1.1}
    \resizebox{\columnwidth}{!}{%
        \begin{tabular}{lcccc}
            \hline
            Model & PLCC & SROCC & KROCC & Overall Score \\
            \hline
            O3 Scores                & 0.6863 & 0.6854 & 0.5127 & 1.8844 \\
            O3-34 Image Scores       & 0.7399 & 0.7255 & 0.5503 & 2.0156 \\
            O3-Metadata Scores       & 0.8137 & 0.8051 & 0.6347 & 2.2535 \\
            O3-Error-Feedback Scores & 0.8122 & 0.8180 & 0.6347 & \textbf{2.2649} \\
            \hline
        \end{tabular}
    }
\end{table}

When the results were examined based on the O3 model, the score performance, which started at 1.88 was increased to the 2.25–2.26 range with the improved methods and metadata. This indicates that providing additional metadata to LMMs significantly increased performance.

\begin{figure}[H]
    \centering
    \includegraphics[width=0.50\textwidth]{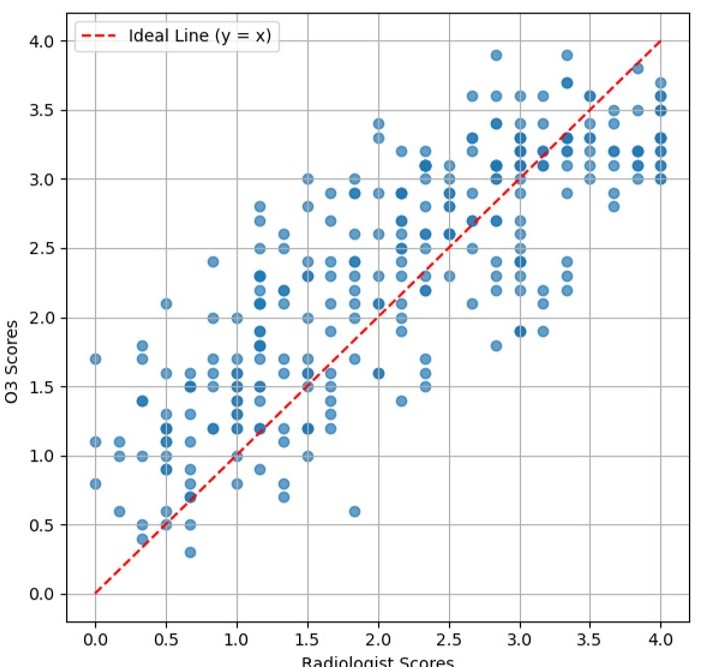}
    \caption{Scatter Plot Comparing Radiologist Scores with O3-Error Feedback Scores}
    \label{fig:scatter_rl}
\end{figure}

\begin{figure}[H]
    \centering
    \includegraphics[width=0.50\textwidth]{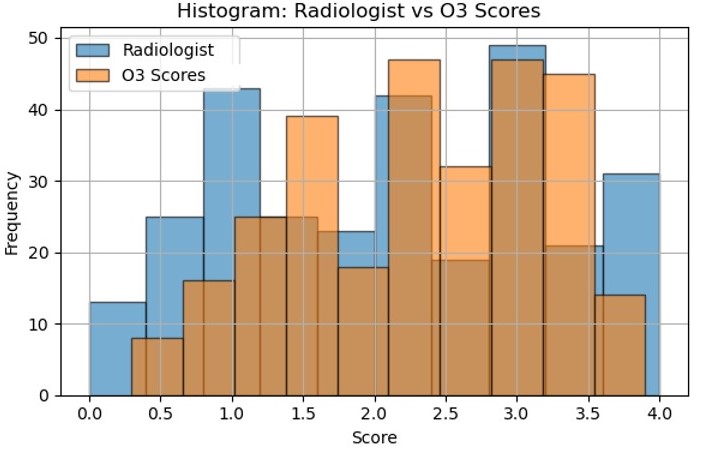}
    \caption{Distribution of Radiologist vs O3-Error Feedback Scores}
    \label{fig:histogram_rl}
\end{figure}
\subsection{Error Feedback–Guided Inference Results}
O3-Error Feedback consistently improved across all correlation metrics, yielding the best LMM-based result in this set of experiments with an Overall Score of 2.2649. The improvement is particularly pronounced in the order-sensitive metrics (SROCC and KROCC), indicating that reasoning patterns selected with reward/punishment can more consistently utilize the Region/Noise context, reducing outlier scores.

O3-Error Feedback demonstrated a strong correlation with radiologist scores, as confirmed by both shown in Fig.~\ref{fig:scatter_rl} and Fig.~\ref{fig:histogram_rl}.

Distribution between radiologist scores and O3-Error Feedback scores shown in Fig.~\ref{fig:scatter_rl}. The majority of points cluster near the ideal line ($y = x$), indicating a strong correlation between model scores and radiologist scores. The agreement is particularly good in the medium- and high-quality regions (range 2.0–3.5), where the model performs similarly to radiologists. Some deviations appear in the low-quality scores (range 0–1), where the model tends to rate images slightly higher than radiologists. This suggests a more tolerant behavior toward noise and artifacts.

Comparison of the score distributions for radiologists and O3-Error Feedback is shown in Fig.~\ref{fig:histogram_rl}. The radiologist score distribution is broader and more spread out toward the extreme values (0 and 4), while the O3-Error Feedback distribution is concentrated in a narrower band (1–3.5). The two distributions match well, especially in the ranges 1.0–2.0 and 2.5–3.5. However, the model used very low (0–0.5) and very high (4.0) scores less frequently, suggesting that O3-Error Feedback is more cautious about assigning extreme values and generally leans toward mid-range scores.

\section{Conclusion}
A language model (LMM)-based methodology was created in this study for assessing quality in low-dose CT acquisitions that differed from prior deep learning-based NR-IQA methodologies. The deep learning paradigms generated accurate scores, but were limited to the numerical case without fully characterizing the multidimensional aspects of human perception. In this investigation, the LMM's linguistic inference ability was harnessed to address this limitation; the system both generated a score as well as identified quality degrading factors with textual linguistic explanations. In this regard, the new methodology generated both numerical as well as interpretative outputs which is more meaningful in the broad clinical context. Experimental results demonstrated that the model established high correlation with radiologist annotations and maintained its generalization ability to other anatomical regions. Moreover, the accurate discrimination of low-dose CT artifacts, such as noise, blurs, and diminished contrast, informed the investigative implications of practical clinical utility from our method. These results suggest LMM-based IQA systems are not only a substitute to existing DL paradigms, but add a valuable and complementary layer of interpretability to existing and future input-output deep learning based perception models.
In addition, our research paves the way for future studies. More extensive and varied datasets, further modality data, and the potential for LMMs to work interactively with radiologists may yield quality assessment that is better, more reliable, and standardized, and add new insights. In addition, the ability to provide both descriptive scoring and descriptive statements regarding diagnostic quality allows the method to enhance interpretability when integrated into clinical decision support systems, strengthening the balance between patient safety and diagnostic accuracy in low-dose CT imaging.

\footnotesize
\bibliographystyle{IEEEbib}
\bibliography{refs}

@article{wang2024comprehensive,
  title={A comprehensive review of multimodal large language models: Performance and challenges across different tasks},
  author={Wang, Jiaqi and Jiang, Hanqi and Liu, Yiheng and Ma, Chong and Zhang, Xu and Pan, Yi and Liu, Mengyuan and Gu, Peiran and Xia, Sichen and Li, Wenjun and others},
journal = J_ARXIV_PREPRINT_ARXIV_2408_01319,
  year={2024}
}

@article{zhu2024adaptive,
  title={Adaptive image quality assessment via teaching large multimodal model to compare},
  author={Zhu, Hanwei and Wu, Haoning and Li, Yixuan and Zhang, Zicheng and Chen, Baoliang and Zhu, Lingyu and Fang, Yuming and Zhai, Guangtao and Lin, Weisi and Wang, Shiqi},
journal = J_ADVANCES_IN_NEURAL_INFORMATION_PROCESSING_SYSTEMS,
  volume={37},
  pages={32611--32629},
  year={2024}
}

@article{varga2025comparative,
  title={Comparative Evaluation of Multimodal Large Language Models for No-Reference Image Quality Assessment with Authentic Distortions: A Study of OpenAI and Claude. AI Models},
  author={Varga, Domonkos},
journal = J_BIG_DATA_AND_COGNITIVE_COMPUTING,
  volume={9},
  number={5},
  pages={132},
  year={2025},
  publisher={MDPI}
}

@inproceedings{you2024depicting,
  title={Depicting beyond scores: Advancing image quality assessment through multi-modal language models},
  author={You, Zhiyuan and Li, Zheyuan and Gu, Jinjin and Yin, Zhenfei and Xue, Tianfan and Dong, Chao},
booktitle = B_EUROPEAN_CONFERENCE_ON_COMPUTER_VISION,
  pages={259--276},
  year={2024},
  organization={Springer}
}

@article{youenhancing,
  title={Enhancing Descriptive Image Quality Assessment with a Large-scale Multi-modal Dataset},
  author={You, Zhiyuan and Gu, Jinjin and Cai, Xin and Li, Zheyuan and Zhu, Kaiwen and Dong, Chao and Xue, Tianfan}
}

@article{chen2024grounding,
  title={Grounding-iqa: Multimodal language grounding model for image quality assessment},
  author={Chen, Zheng and Zhang, Xun and Li, Wenbo and Pei, Renjing and Song, Fenglong and Min, Xiongkuo and Liu, Xiaohong and Yuan, Xin and Guo, Yong and Zhang, Yulun},
journal = J_ARXIV_PREPRINT_ARXIV_2411_17237,
  year={2024}
}

@article{chen2024iqagpt,
  title={IQAGPT: computed tomography image quality assessment with vision-language and ChatGPT models},
  author={Chen, Zhihao and Hu, Bin and Niu, Chuang and Chen, Tao and Li, Yuxin and Shan, Hongming and Wang, Ge},
journal = J_VISUAL_COMPUTING_FOR_INDUSTRY_BIOMEDICINE_AND_ART,
  volume={7},
  number={1},
  pages={20},
  year={2024},
  publisher={Springer}
}

@article{wu2023q,
  title={Q-align: Teaching lmms for visual scoring via discrete text-defined levels},
  author={Wu, Haoning and Zhang, Zicheng and Zhang, Weixia and Chen, Chaofeng and Liao, Liang and Li, Chunyi and Gao, Yixuan and Wang, Annan and Zhang, Erli and Sun, Wenxiu and others},
journal = J_ARXIV_PREPRINT_ARXIV_2312_17090,
  year={2023}
}

@inproceedings{you2025teaching,
  title={Teaching large language models to regress accurate image quality scores using score distribution},
  author={You, Zhiyuan and Cai, Xin and Gu, Jinjin and Xue, Tianfan and Dong, Chao},
booktitle = B_PROCEEDINGS_OF_THE_COMPUTER_VISION_AND_PATTERN_RECOGNITION_CON,
  pages={14483--14494},
  year={2025}
}

@article{lee2025low,
  title={Low-dose computed tomography perceptual image quality assessment},
  author={Lee, Wonkyeong and Wagner, Fabian and Galdran, Adrian and Shi, Yongyi and Xia, Wenjun and Wang, Ge and Mou, Xuanqin and Ahamed, Md Atik and Imran, Abdullah Al Zubaer and Oh, Ji Eun and others},
journal = J_MEDICAL_IMAGE_ANALYSIS,
  volume={99},
  pages={103343},
  year={2025},
  publisher={Elsevier}
}

@inproceedings{zhu2020metaiqa,
  title={MetaIQA: Deep meta-learning for no-reference image quality assessment},
  author={Zhu, Hancheng and Li, Leida and Wu, Jinjian and Dong, Weisheng and Shi, Guangming},
booktitle = B_PROCEEDINGS_OF_THE_IEEE_CVF_CONFERENCE_ON_COMPUTER_VISION_AND,
  pages={14143--14152},
  year={2020}
}

@article{bosse2017deep,
  title={Deep neural networks for no-reference and full-reference image quality assessment},
  author={Bosse, Sebastian and Maniry, Dominique and M{\"u}ller, Klaus-Robert and Wiegand, Thomas and Samek, Wojciech},
journal = J_IEEE_TRANSACTIONS_ON_IMAGE_PROCESSING,
  volume={27},
  number={1},
  pages={206--219},
  year={2017},
  publisher={IEEE}
}

@article{madhusudana2022image,
  title={Image quality assessment using contrastive learning},
  author={Madhusudana, Pavan C and Birkbeck, Neil and Wang, Yilin and Adsumilli, Balu and Bovik, Alan C},
journal = J_IEEE_TRANSACTIONS_ON_IMAGE_PROCESSING_2,
  volume={31},
  pages={4149--4161},
  year={2022},
  publisher={IEEE}
}

@inproceedings{yang2022maniqa,
  title={Maniqa: Multi-dimension attention network for no-reference image quality assessment},
  author={Yang, Sidi and Wu, Tianhe and Shi, Shuwei and Lao, Shanshan and Gong, Yuan and Cao, Mingdeng and Wang, Jiahao and Yang, Yujiu},
booktitle = B_PROCEEDINGS_OF_THE_IEEE_CVF_CONFERENCE_ON_COMPUTER_VISION_AND,
  pages={1191--1200},
  year={2022}
}

@article{talebi2018nima,
  title={NIMA: Neural image assessment},
  author={Talebi, Hossein and Milanfar, Peyman},
journal = J_IEEE_TRANSACTIONS_ON_IMAGE_PROCESSING_3,
  volume={27},
  number={8},
  pages={3998--4011},
  year={2018},
  publisher={IEEE}
}

@inproceedings{byun2021fbi,
  title={Fbi-denoiser: Fast blind image denoiser for poisson-gaussian noise},
  author={Byun, Jaeseok and Cha, Sungmin and Moon, Taesup},
booktitle = B_PROCEEDINGS_OF_THE_IEEE_CVF_CONFERENCE_ON_COMPUTER_VISION_AND_2,
  pages={5768--5777},
  year={2021}
}

\end{document}